\documentclass{article}
\usepackage[bookmarks=false,hyperfigures=false,hidelinks=true]{hyperref}
\usepackage{spconf,amsmath,amssymb,bbm,mathtools,subfigure,doi,xspace}
\usepackage{url}
\usepackage{diagbox}
\usepackage{cite}
\usepackage{ifpdf}

\ifpdf
  \usepackage{graphicx,color}
  \DeclareGraphicsExtensions{.pdf,.png}
\else
  \usepackage{graphicx,color}
  \DeclareGraphicsExtensions{.pstex,.ps,.eps,.png}
\fi

\newcommand{\ie}{i.e.\xspace}

\newcommand{\eq}[1]{(\ref{eq:#1})}
\newcommand{\fig}[1]{Fig.~\ref{fig:#1}}

\newcommand{\tab}[1]{Table~\ref{tab:#1}}
\newcommand{\sctn}[1]{Sec.~\ref{sec:#1}}

\newcommand{\mb}[1]{\mathbf{#1}}

\newcommand{\mbb}[1]{\mathbb{#1}}

\newcommand{\sign}{\mathop{\mathrm{sign}}}

\newcommand{\norm}[1]{\left\| #1 \right\|}
\newcommand{\abs}[1]{\left| #1 \right|}

\DeclareMathOperator*{\argmin}{arg\,min}
\DeclarePairedDelimiterX{\normsz}[1]{\lVert}{\rVert}{#1}
\DeclarePairedDelimiterX{\parensz}[1]{(}{)}{#1}

\newcommand{\resizemath}[2]{%
  \resizebox{#1}{!}{%
    $ \displaystyle #2 $%
  }%
}%

\ninept

\allowdisplaybreaks[3]

\title{Convolutional Sparse Coding with Overlapping Group Norms}

\name{Brendt Wohlberg\thanks{This research was supported by the
    U.S. Department of Energy through the LANL/LDRD Program.}}
\address{Theoretical Division\\
  Los Alamos National Laboratory\\
  Los Alamos, NM 87545, USA}

\begin{document}

\maketitle

\begin{abstract}
 The most widely used form of convolutional sparse coding uses an $\ell_1$ regularization term. While this approach has been successful in a variety of applications, a limitation of the $\ell_1$ penalty is that it is homogeneous across the spatial and filter index dimensions of the sparse representation array, so that sparsity cannot be separately controlled across these dimensions.
The present paper considers the consequences of replacing the $\ell_1$ penalty with a mixed group norm, motivated by recent theoretical results for convolutional sparse representations. Algorithms are developed for solving the resulting problems, which are quite challenging, and the impact on the performance of the denoising problem is evaluated. The mixed group norms are found to perform very poorly in this application. While their performance is greatly improved by introducing a weighting strategy, such a strategy also improves the performance obtained from the much simpler and computationally cheaper $\ell_1$ norm.
\end{abstract}


\section{Introduction}
\label{sec:intro}

A standard sparse representation of a signal $\mb{s}$ is a linear representation of the form $D \mb{x} \approx \mb{s}$, where $D$ is the \emph{dictionary} and $\mb{x}$ is the coefficient vector. A \emph{convolutional sparse representation}~\cite{zeiler-2010-deconvolutional}\cite[Sec. II]{wohlberg-2016-efficient} replaces this form with a sum of convolutions $\sum_m \mb{d}_m \ast \mb{x}_m \approx \mb{s}$, where the elements of the dictionary $\mb{d}_m$ are linear filters, and the representation consists of the stack of coefficient maps $\mb{x}_m$, each of which is the same size as $\mb{s}$. Recently there has been a significant growth in interest in the use of this type of representation for problems in signal and image processing~\cite{gu-2015-convolutional, jao-2015-informed, cogliati-2015-piano, liu-2016-image, wohlberg-2016-convolutional2, zhang-2016-convolutional, quan-2016-compressed, zhang-2017-convolutional}. While convolutional sparse representations have been found to provide state of the art performance in a variety of image reconstruction problems, a notable and perhaps surprising omission is denoising subject to Gaussian white noise. While this is arguably the simplest of all image reconstruct problems, no competitive convolutional sparse representation based solution for this problem has been reported in the literature, and there is some evidence that the highly overcomplete nature of the representation leaves it at a disadvantage with respect to the more traditional sparsity-based denoising techniques~\cite{carrera-2017-sparse2}.

Most of the signal and image processing applications listed above have used the same form of the convolutional sparse coding (CSC) problem, with an $\ell_1$ regularization term on the stack of coefficient maps. While simple $\ell_1$ regularization has provided good performance in these applications, it does not exploit the rich structure present in the convolutional representation, which typically exhibits patterns of support corresponding to the edge structure of the image being represented. Furthermore, as pointed out in~\cite{luo-2016-laplacian}, the $\ell_1$ norm of a multi-dimensional array is homogeneous on all dimensions, with no way to separately penalise sparsity across the spatial dimensions and down the filter index dimension of the stack of coefficient maps. There has been some work on applying different or additional regularization terms to the CSC problem~\cite{szlam-2011-structured,
luo-2016-laplacian, wohlberg-2016-convolutional, wohlberg-2017-convolutional, cogliati-2017-piano, zhang-2017-convolutional}, but these approaches tend to be specialised to specific problems, or have other limitations, none of them providing a suitable generic replacement for the $\ell_1$ norm.

The primary purpose of the present paper is to consider mixed group norms as a potential replacement for the $\ell_1$ norm in CSC problems. This is motivated by the ability of these norms to impose useful forms of structured sparsity, which has been widely exploited in machine learning (see e.g.~\cite{zhou-2010-exclusive}), as well as by a recent theoretical work~\cite{papayan-2017-working1} arguing that the $\ell_{0,\infty}$ ``norm'' is appropriate when working with convolutional representations, and speculating that its convex relaxation, the $\ell_{1,\infty}$ norm, may offer performance advantages in practice~\cite[Sec. VI]{papayan-2017-working2}. We show that this expectation is not realized: for the classical white-noise denoising problem, at least, mixed group norms are significantly outperformed by the computationally cheaper $\ell_1$ norm regularization.

The outline of the remainder of this paper is as follows. \sctn{cbpdn} introduces the most common form of CSC problem and an efficient Alternating Direction Method of Multipliers (ADMM) algorithm for solving it. \sctn{mixnorm} defines the mixed norms that will be considered, and explains why the appropriate way of applying them for convolutional sparse representations necessitates the use of overlapping groups. Sec. \ref{sec:algl1inf} and \ref{sec:l12} propose algorithms for solving CSC problems with $\ell_{1,\infty}$ and $\ell_{1,2}$ mixed norms with overlapping groups. These algorithms are described in some detail since these are difficult problems for which there are no existing algorithms in the literature; although these forms of the CSC problem do not provide good performance in the denoising problem in which they are tested, the algorithms are expected to be of value to other researchers wishing to further explore the properties of these problems, motivated by the theoretical results discussed above. \sctn{results} provides the denoising performance comparisons between these CSC methods, the standard CSC problem with an $\ell_1$ norm, and a standard block-based (non-convolutional) sparse coding denoising method. \sctn{weighting} discusses the reasons for the poor performance of the mixed-norm CSC methods, and shows that this performance can be substantially improved by a suitable weighting strategy. A weighting strategy for the standard $\ell_1$ CSC method is also proposed in~\sctn{l1weighting}, and found to not only outperform the mixed-norm CSC methods, but to also provide competitive performance with the block-based sparse coding reference method. Finally, conclusions are presented in \sctn{cnclsn}.

\section{Convolutional Sparse Coding}
\label{sec:cbpdn}

The $\ell_1$ penalised form of convolutional sparse coding is
\vspace{-1mm}
\begin{equation}
\argmin_{\{\mb{x}_m\}} \frac{1}{2} \normsz[\Big]{\sum_m \mb{d}_m \ast \mb{x}_m
- \mb{s}}_2^2 + \lambda \sum_m \norm{\mb{w}_m \odot \mb{x}_m}_1 \; ,
\label{eq:convbpdn}
\vspace{-1mm}
\end{equation}
where the $\mb{w}_m$ allow weighting of the $\ell_1$ term.  At present, the most efficient approach to solving this problem~\cite{wohlberg-2016-efficient} is via the Alternating Direction Method of Multipliers (ADMM)~\cite{boyd-2010-distributed} framework. An outline of this method is presented here as a reference for the modifications proposed in following sections.

Problem ~\eq{convbpdn} can be written as
\vspace{-1mm}
\begin{equation}
\argmin_{\mb{x}} \frac{1}{2} \normsz[\big]{D \mb{x}
- \mb{s}}_2^2 + \lambda \norm{\mb{w} \odot \mb{x}}_1 \; ,
\label{eq:convbpdnblk}
\end{equation}
where $D_m$ is a linear operator such that $D_m \mb{x}_m = \mb{d}_m \ast \mb{x}_m$, and $D$, $\mb{w}$, and $\mb{x}$ are the block matrices/vectors
\begin{equation}
\resizemath{.75\hsize}{
D = \left( \begin{array}{ccc}D_0 & D_1 &
    \ldots \end{array} \right)  \;\;
\mb{w} = \left( \begin{array}{c}  \mb{w}_0\\ \mb{w}_1\\
    \vdots  \end{array} \right)
\;\;
\mb{x} = \left( \begin{array}{c}  \mb{x}_0\\ \mb{x}_1\\
    \vdots  \end{array} \right) \;.
}
\vspace{-1mm}
\label{eq:dxcbpdn}
\end{equation}
This problem can be expressed in ADMM standard form as
\vspace{-1mm}
\begin{equation}
\argmin_{\mb{x},\mb{y}} \frac{1}{2} \normsz[\big]{
    D \mb{x} - \mb{s}}_2^2 \!+\! \lambda
  \norm{\mb{w} \odot \mb{y}}_1 \text{ s.t. } \mb{x} =
  \mb{y}  \;\; ,
\label{eq:cbpdnsplit}
\vspace{-1mm}
\end{equation}
which can be solved via the ADMM iterations
\setlength{\abovedisplayskip}{6pt}
\setlength{\belowdisplayskip}{\abovedisplayskip}
\setlength{\abovedisplayshortskip}{0pt}
\setlength{\belowdisplayshortskip}{3pt}
\begin{align}
  \mb{x}^{(j+1)} &= \argmin_{\mb{x}} \frac{1}{2}
  \normsz[\big]{D \mb{x} - \mb{s}}_2^2 + %
  \frac{\rho}{2}  \norm{
    \mb{x} - \mb{y}^{(j)} + \mb{u}^{(j)}}_2^2 \label{eq:bpdnxprob} \\
  \mb{y}^{(j+1)} &= \argmin_{\mb{y}} \lambda
  \norm{\mb{w} \odot \mb{y}}_1 + %
   \frac{\rho}{2}
 \norm{ \mb{x}^{(j+1)} - \mb{y} +
    \mb{u}^{(j)}}_2^2 \label{eq:bpdnyprob}  \\
  \mb{u}^{(j+1)} &= \mb{u}^{(j)} + \mb{x}^{(j+1)} -
  \mb{y}^{(j+1)} \; . \label{eq:bpdnuprob}
\vspace{-1mm}
\end{align}

The only computationally expensive step is~\eq{bpdnxprob}, which involves solving the linear system
\begin{equation}
(D^T D + \rho I) \mb{x} = D^H \mb{s} + \rho \left(\mb{y} - \mb{u} \right) \; .
\end{equation}
This very large linear system can be solved efficiently by exploiting the Sherman-Morrison formula in the DFT domain~\cite{wohlberg-2014-efficient}.

\section{Mixed Norms with Overlapping Groups}
\label{sec:mixnorm}

The $\ell_{p,q}$ mixed matrix norm~\cite{kowalski-2009-sparse} is
$
  \|X\|_{p,q}  = \parensz[\big]{ \sum_i \| \mb{x}_i \|_p^q }^{1/q}
$,
where $\mb{x}_i$ is row $i$ of matrix $X$. (Note that some authors use a notation that reverses the roles of $p$ and $q$.) A special case is
$
  \|X\|_{p,\infty}  = \max_i \left( \| \mb{x}_i \|_p \right)
$.
These mixed norms can also be defined in the context of arbitrary \emph{groups} of coefficients in a vector. If we define $g_i(\mb{x})$ as the function extracting the elements within the $i^{\text{th}}$ group of $\mb{x}$, the corresponding $\ell_{p,q}$ group norm of $\mb{x}$ is
\begin{equation}
  \norm{\mb{x}}_{p,q}  = \parensz[\bigg]{ \sum_i \norm{ g_i(\mb{x}) }_p^q}^{1/q} \;\;,
\end{equation}
and the $\ell_{p,\infty}$ norm is
$
  \norm{\mb{x}}_{p,\infty}  = \max_i \parensz[\big]{ \norm{ g_i(\mb{x}) }_p }
$.
The most widely used mixed norm is the $\ell_{2,1}$ norm, which promotes group sparsity in the sense that only a few groups are active, but the representation is not sparse within each active group. Solving problems involving this norm becomes challenging when the groups overlap~\cite[Sec. 6.4.2]{boyd-2010-distributed}, the leading strategy for dealing with such cases involving the use of variable duplication strategies~\cite[Sec. 6.4.2]{boyd-2010-distributed}\cite[Sec. 3.1]{deng-2013-group}

We will consider CSC with the $\ell_{1,\infty}$ norm, as proposed in~\cite[Sec. VI]{papayan-2017-working2}, as well as with the $\ell_{1,2}$ norm, which has been shown to have useful properties in signal processing applications~\cite{kowalski-2009-sparsity}. It may seem that the most straightforward way of applying these norms to the CSC coefficient map stack is to compute the $\ell_1$ norm along the filter index at each spatial location (i.e. non-overlapping groups consisting of all coefficients at the same spatial location), and the $\ell_{\infty}$ or $\ell_2$ norm on the resulting sum, but such an approach is problematic since it completely ignores the potentially different spatial properties of the different filters in the dictionary. For the mixed group norms to function in a coherent way, the groups should consist of sets of filters that affect the same spatial location in the image being represented, but it is quite possible that two nominally-aligned filters have their centres of mass in different positions, so that the corresponding coefficients at the same spatial indices in the stack of coefficient maps affect different spatial locations in the image.

\begin{figure}[htbp]
  \centering
  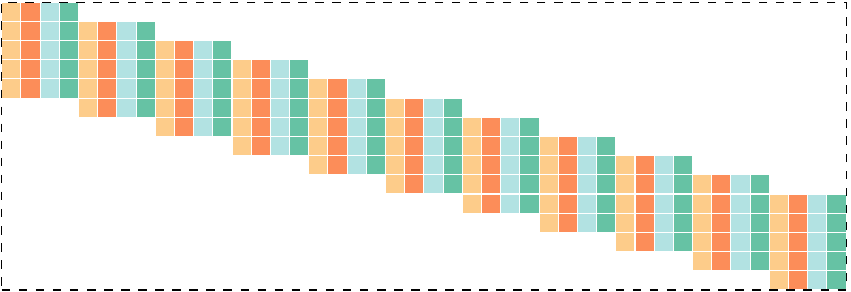
  \caption{A convolutional dictionary can be viewed as a structured dictionary for an entire signal, constructed from all translations of a smaller block dictionary. In this example the block dictionary consists of 4 atoms, each in $\mbb{R}^5$, and the usual circular boundary conditions are not depicted for simplicity.
}
  \label{fig:dctglbl}
\end{figure}

\begin{figure}[htbp]
  \centering
  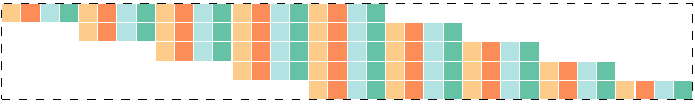
  \caption{The smallest part of the signal dictionary that captures all of its properties is a horizontal stripe, of the same height as the generating block dictionary.
The coefficients corresponding to each stripe constitute a coefficient group.
}
  \label{fig:dctptch}
\end{figure}

\begin{figure}[htbp]
   \centering
   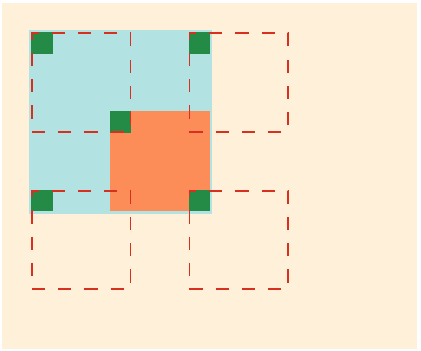
   \caption{Spatial arrangement of coefficient groups in an image. The large beige rectangle represents part of the spatial support of image and coefficient maps. The central green square represents the coefficient corresponding to the filter position in the image indicated in orange, and the four outlying green squares represent coefficients corresponding to the filter placements indicated by dashed red lines, which are the furthest filter placements that still overlap with the central one. The light blue square indicates the set of all coefficients that correspond to filter placements that overlap the central orange patch, \ie a single coefficient group (including all coefficients within this square across all coefficient maps).}
   \label{fig:grp}
 \end{figure}

To understand the correct notion of a coefficient group, we need to consider the view of a convolutional dictionary as a structured dictionary for the entire signal, constructed from all possible translations of a smaller block dictionary (the set of filters in the convolutional dictionary stacked as the columns of a matrix), as depicted in~\fig{dctglbl}. As pointed out in~\cite{papayan-2017-working1}, the smallest part of this dictionary that captures all of its important properties is a horizontal stripe across it, as depicted in~\fig{dctptch}. Each of these  stripes consists of all dictionary atoms that contribute to the reconstruction of a signal patch with the same ``height'' as the stripe, and at the same location. It is the set of coefficients corresponding to each stripe that define the groups on which the the $\ell_{0,\infty}$  ``norm'' and $\ell_{1,\infty}$ norm proposed in~\cite{papayan-2017-working1} is based, and which will be used here.
The spatial arrangement of one of these groups in the coefficient maps for an image is illustrated in~\fig{grp}.

Since these groups are highly overlapping we might expect, considering the difficulty of dealing with $\ell_{2,1}$ norm with overlapping groups, that solving problems involving the $\ell_{1,\infty}$ and $\ell_{1,2}$ norms with overlapping groups would pose considerable difficulties. The variable replication strategy that has been widely applied for the overlapping-group $\ell_{2,1}$ norm is not a viable option here due to the very large number of groups, which would greatly expand the memory requirements of the convolutional representation, which are already high. It turns out, however, that there is a much cheaper variant of the variable duplication strategy that can be applied for $\ell_{1,q}$ problems --- since the inner norm is a sum of absolute values, we need only replicate the sum of each group, rather than the entire group. Efficient computation of these group sums, which is essential since the coefficient array can be very large, can be achieved by convolving with a suitable kernel of unit entries to compute the spatial sums, which are then summed over the filter index.

\section{Algorithms for the $\ell_{1,\infty}$ Norm}
\label{sec:algl1inf}

The CSC $\ell_{1,\infty}$ problem can be written as
\begin{equation}
  \argmin_{\mb{x}}  \; (1/2) \norm{ D \mb{x}  - \mb{s} }_2^2 + \lambda
  \norm{\mb{x}}_{1,\infty} \;\;,
  \label{eq:cbpdnl1infty}
\end{equation}
where $\norm{\mb{x}}_{1,\infty} = \max_i \left( \norm{ g_i(\mb{x}) }_1 \right)$, with the $g_i(\cdot)$ defined by the group structure discussed above.
Define $G_m$ such that $G_m \mb{x}_m = \mb{1}_m \ast \mb{x}_m$,
where $\mb{1}_m$ is a unit filter of the same support as $\mb{d}_m$, and
$  G = \left( \begin{array}{ccc}G_0 & G_1 &
      \ldots \end{array} \right)$,
allowing us to express $\norm{\mb{x}}_{1,\infty}$ as $\max ( G \abs{\mb{x}} )$, where $G \abs{\mb{x}}$ can be efficiently computed in the DFT domain since it is defined in terms of a set of convolutions. We consider two different approaches to solving the resulting problem.

\subsection{Nested ADMM Algorithms}
\label{sec:proxopl1inf}

We start by expressing~\eq{cbpdnl1infty} in ADMM form, as
\begin{equation}
  \argmin_{\mb{x}, \mb{y}}  \; (1/2) \norm{ D \mb{x}  - \mb{s} }_2^2 + \lambda
  \max ( G \abs{\mb{y}} )  \;\; \text{s.t.} \;\; \mb{x}
  = \mb{y} \;\;.
  \label{eq:cbpdnl1inftyadmm}
\end{equation}
The corresponding ADMM iterations are
\begin{align}
  \mb{x}^{(k+1)} &= \argmin_{\mb{x}} \frac{1}{2}\norm{D \mb{x} -
                   \mb{s} }_2^2 + \frac{\rho}{2} \norm{ \mb{x} - \mb{y}^{(k)} +
                   \mb{u}^{(k)}}_2^2 \label{eq:cbpdnl1inftyx} \\
  \mb{y}^{(k+1)} &= \argmin_{\mb{y}} \lambda \max ( G \abs{\mb{y}} ) +
                   \frac{\rho}{2} \norm{ \mb{x}^{(k+1)} \!-\! \mb{y} \!+\!
                   \mb{u}^{(k)}}_2^2  \label{eq:cbpdnl1inftyy} \\
  \mb{u}^{(k+1)} &= \mb{u}^{(k)} + \mb{x}^{(k+1)} -
                   \mb{y}^{(k+1)} \;\; .
\end{align}
Subproblem~\eq{cbpdnl1inftyx} can be solved via the standard DFT-domain Sherman-Morrison approach~\cite{wohlberg-2014-efficient}, but solving subproblem~\eq{cbpdnl1inftyy} involves computing the proximal operator~\cite{parikh-2014-proximal} of $\lambda \max ( G \abs{\mb{x}} )$, which we turn to now.

The proximal operator of $\lambda  \max ( G \abs{\mb{x}} )$ is
\begin{equation}
\argmin_{\mb{x}} \lambda \max ( G \abs{\mb{x}} ) + (1/2) \norm{\mb{x} - \mb{v}}_2^2 \;\;,
\label{eq:mxgprox}
\end{equation}
where we overload symbols $\lambda$ and $\mb{x}$ to avoid dealing with a profusion of symbols.  Now, since the term $\lambda \max ( G \abs{\mb{x}} )$ depends only on the absolute value of $\mb{x}$, it is clear that the sign of the solution will be the same as that of $\mb{v}$; if the sign differs on any coordinate, we can find a lower cost solution by switching the sign to match that of $\mb{v}$, reducing the cost of the $\ell_2$ term and leaving the other term invariant. We can therefore project the problem to the positive orthant (i.e., solve for the proximal operator at $\abs{\mb{v}}$ instead of $\mb{v}$) and then recover the signed solution by point-wise multiplication of the solution on the positive orthant by $\sign(\mb{v})$. This allows problem~\eq{mxgprox} to be further simplified to
\begin{equation}
\argmin_{\mb{x}} \lambda \max ( G \mb{x} ) + \frac{1}{2} \norm{\mb{x} - \mb{v}}_2^2 \quad \text{ s.t. } \quad \mb{x} \geq 0 \;\;.
\label{eq:mxgproxnn}
\end{equation}

Writing in ADMM form we have
\begin{align}
\argmin_{\mb{x}, \mb{y}_0, \mb{y}_1} & \;
\frac{1}{2} \norm{\mb{x} - \mb{v}}_2^2 + \lambda \max ( \mb{y}_0 )
+ \iota_{NN}(\mb{y}_1) \nonumber \\ \;\; & \text{s.t.} \;\;
\alpha_0 \mb{y}_0 = \alpha_0 G \mb{x} \;\; \alpha_1 \mb{y}_1 = \alpha_1 \mb{x} \;\;,
\label{eq:prxgmxsplit}
\end{align}
where $\iota_{NN}(\cdot)$ is the indicator function of the non-negativity constraint, and scalars $\alpha_0, \alpha_1$ are introduced to allow compensation for the potentially large imbalance in the magnitudes of the Augmented Lagrangian~\cite{boyd-2010-distributed} terms corresponding to variables $\mb{y}_0$ and $\mb{y}_1$. (It turns out that reliable convergence of the algorithms depends on selecting suitable values for $\alpha_0, \alpha_1$.)
The corresponding ADMM iterations are
\begin{align}
  \mb{x}^{(k+1)} &= \argmin_{\mb{x}} \; \frac{1}{2}\norm{ \mb{x} -
                   \mb{v} }_2^2 + \frac{\rho}{2} \norm{ \alpha_0 G \mb{x} -
                   \alpha_0 \mb{y}_0^{(k)} + \mb{u}_0^{(k)}}_2^2 \nonumber \\
                & \qquad \qquad + \frac{\rho}{2} \norm{ \alpha_1 \mb{x} -
                   \alpha_1 \mb{y}_1^{(k)} + \mb{u}_1^{(k)}}_2^2 \label{eq:prxx}
                   \\
  \mb{y}_0^{(k+1)} &= \argmin_{\mb{y}_0} \lambda \max ( \mb{y}_0 ) +
                   \frac{\rho}{2} \norm{ \alpha_0 G \mb{x}^{(k+1)} \!-
                   \alpha_0 \mb{y}_0 +
                   \mb{u}_0^{(k)}}_2^2 \nonumber %
                   \\
  \mb{y}_1^{(k+1)} &= \argmin_{\mb{y}_1} \iota_{NN}(\mb{y}_1)  +
                   \frac{\rho}{2} \norm{ \alpha_1 \mb{x}^{(k+1)} \!-
                   \alpha_1 \mb{y}_1 +
                   \mb{u}_1^{(k)}}_2^2 \nonumber  %
                   \\
  \mb{u}_0^{(k+1)} &= \mb{u}_0^{(k)} + \alpha_0 G \mb{x}^{(k+1)} -
                   \alpha_0 \mb{y}_0^{(k+1)} \nonumber \\
 \mb{u}_1^{(k+1)} &= \mb{u}_1^{(k)} + \alpha_1 \mb{x}^{(k+1)} -
                   \alpha_1 \mb{y}_1^{(k+1)} \nonumber \;\; .
\end{align}
Solving~\eq{prxx} involves solving the linear system
\begin{align}
\parensz[\bigg]{G^T G + \alpha_0^{-2} (\alpha_1^2 + \rho^{-1}) I} \mb{x} =& \alpha_0^{-2} \rho^{-1} \mb{v} + G^T(\mb{y}_0 - \alpha_0^{-1}\mb{u}_0) + \nonumber \\ &  \alpha_0^{-2} \alpha_1^2 (\mb{y}_1 - \alpha_1^{-1} \mb{u}_1) \;,
\end{align}
which can be solved in the DFT domain via the Sherman-Morrison method~\cite{wohlberg-2014-efficient}, and the $\mb{y}_0$ and $\mb{y}_1$ subproblems can be solved via the proximal operators of the max function (see Sec. 6.4.1 and 6.5.2 in ~\cite{parikh-2014-proximal}) and the non-negativity constraint (clipping to zero) respectively.

The algorithm for solving~\eq{cbpdnl1inftyadmm} converges reliably if the proximal operator subproblem~\eq{mxgprox} is solved to sufficient accuracy, but tends to be slow due to the nested iterations -- problem~\eq{cbpdnl1infty}
is solved via an iterative algorithm, and step~\eq{cbpdnl1inftyy} of this iterative algorithm is itself solved via an iterative algorithm. This cost can be mitigated by (i) performing a warm start of the inner optimization at each outer iteration, and (ii) careful selection of a stopping criterion for the inner problem based on the relative change in functional value. Nevertheless, it remains very high compared with the standard problem~\eq{convbpdn}: on a test problem with an $8 \times 8 \times 128$ dictionary and $128 \times 128$ image, for which 250 iterations of the ADMM algorithm for~\eq{convbpdn} took 262 seconds to complete, this algorithm required 844 seconds to complete the same number of iterations. It also requires tuning of quite a large number of parameters, including $\alpha_0$ and $\alpha_1$, the penalty parameters $\rho$ for both outer and inner ADMM algorithms, and the stopping tolerance for the inner iterations.

\subsection{Mapping to a Non-Negative Problem}
\label{sec:mapnonegl1inf}

Given the cost of the previous algorithm, we consider an alternative approach.
We start by solving a variant of~\eq{cbpdnl1infty} including a
non-negativity constraint
\begin{equation}
  \argmin_{\mb{x}}  \; (1/2) \norm{ D \mb{x}  - \mb{s} }_2^2 + \lambda
  \max ( G \abs{\mb{x}} ) \;\; \text{s.t.} \;\; \mb{x} \geq 0  \;\;,
  \label{eq:cbpdnl1inftynn}
\end{equation}
which we can pose in ADMM form as
\begin{align}
  \argmin_{\mb{x}, \mb{y}_0, \mb{y}_1}  \; & (1/2) \norm{ D \mb{x}  - \mb{s} }_2^2 + \lambda
  \max ( \mb{y_0} ) + \iota_{NN}(\mb{y_1}) \nonumber \\
  \;\; & \text{s.t.} \;\; \alpha_0 \mb{y}_0
  = \alpha_0 G \mb{x}  \quad \alpha_0 \mb{y}_1 = \alpha_0 \mb{x}  \;\;,
  \label{eq:cbpdnl1inftynnadmm}
\end{align}
where scalars $\alpha_0, \alpha_0$ are included for the same reason as before.
The corresponding ADMM iterations are
\begin{align}
  \mb{x}^{(k\!+\!1)} = & \argmin_{\mb{x}} \frac{1}{2}\norm{D \mb{x} -
                     \mb{s} }_2^2 + \frac{\rho}{2} \norm{ \alpha_0 G \mb{x}
          \!-\! \alpha_0 \mb{y}_0^{(k)} \!+\! \mb{u}_0^{(k)}}_2^2 + \nonumber \\
                   &  \frac{\rho}{2} \norm{ \alpha_1 \mb{x}
                     - \alpha_1 \mb{y}_1^{(k)} + \mb{u}_1^{(k)}}_2^2
                     \label{eq:cbpdngmxnnx} \\
  \mb{y}_0^{(k\!+\!1)} = & \argmin_{\mb{y}_0} \lambda \! \max ( \mb{y}_0 )
                       \!+\! \frac{\rho}{2} \norm{ \alpha_0 G \mb{x}^{(k)}
                       \!-\! \alpha_0 \mb{y}_0 \!+\! \mb{u}_0^{(k)}}_2^2 \!\!
                         \label{eq:cbpdngmxnny0} \\
  \mb{y}_1^{(k\!+\!1)} = & \argmin_{\mb{y}_1}
                       \iota_{NN}(\mb{y_1}) +
                       \frac{\rho}{2} \norm{ \alpha_1 \mb{x}^{(k)}
                       \!-\! \alpha_1 \mb{y}_1 \!+\! \mb{u}_1^{(k)}}_2^2
                         \label{eq:cbpdngmxnny1} \\
  \mb{u}_0^{(k\!+\!1)} = & \; \mb{u}_0^{(k)} + \alpha_0 G \mb{x}^{(k\!+\!1)} -
                       \alpha_0 \mb{y}_0^{(k\!+\!1)} \\
  \mb{u}_1^{(k\!+\!1)} = & \; \mb{u}_1^{(k)} + \alpha_1 \mb{x}^{(k+1)} -
                       \alpha_1 \mb{y}_1^{(k+1)} \;\; .
\end{align}

Subproblem~\eq{cbpdngmxnnx} involves solving the linear system
\begin{align}
(D^TD + \rho \alpha_0^2 G^TG + \rho \alpha_1^2 I) \mb{x} = & D^T \mb{s} + \nonumber \\ \rho \alpha_0 G^T (\alpha_0 \mb{y}_0^{(k)}  - \mb{u}_0^{(k)}) +
  & \rho \alpha_1 (\alpha_1 \mb{y}_1^{(k)} - \mb{u}_1^{(k)}) \;,
\end{align}
which can be efficiently solved in the DFT domain by iterated application of the Sherman-Morrison formula~\cite[Appendix D]{wohlberg-2016-efficient}. Subproblems~\eq{cbpdngmxnny0} and~\eq{cbpdngmxnny1} can be solved via the proximal operators of of the max function (see Sec. 6.4.1 and 6.5.2 in ~\cite{parikh-2014-proximal}) and the non-negativity constraint (clipping to zero) respectively.

Finally, in order to work around the non-negativity constraint, we replace $D \mb{x}$ and $G \mb{x}$ with
\[
(\begin{array}{cc} D & -D \end{array}) \left( \begin{array}{c} \mb{x}_0 \\ \mb{x}_1 \end{array}  \right) \;\; \text{and} \;\;
 (\begin{array}{cc} G & G \end{array}) \left( \begin{array}{c} \mb{x}_0 \\ \mb{x}_1 \end{array}  \right)
\]
respectively, allowing us to recover the solution of the original %
problem~\eq{cbpdnl1infty},
without the non-negativity constraint, as $\mb{x}_0 - \mb{x}_1$.

The outer iterations of this algorithm are substantially faster than those of~\sctn{proxopl1inf}, taking 415 seconds to complete the same test problem for which that algorithm required 844 seconds (as reported in ~\sctn{proxopl1inf}). Although some care is necessary in choosing $\rho$, $\alpha_0$ and $\alpha_1$, it has fewer parameters to tune than the algorithm of~\sctn{proxopl1inf}, and can be made to converge reliably.

\section{Algorithms for the $\ell_{1,2}$ Norm}
\label{sec:l12}

The CSC $\ell_{1,2}$ problem can be written as
\begin{equation}
  \argmin_{\mb{x}}  \; (1/2) \norm{ D \mb{x}  - \mb{s} }_2^2 + \lambda
  \norm{\mb{x}}_{1,2} \;\;,
  \label{eq:cbpdnl12}
\end{equation}
where $\norm{\mb{x}}_{1,2} = \sqrt{ \sum_i  \norm{ g_i(\mb{x}) }_1^2 }$, with the $g_i(\cdot)$ defined by the group structure discussed in~\sctn{mixnorm}. Using $G$ as defined in~\sctn{algl1inf}, we can express $\norm{\mb{x}}_{1,2}$ as $\norm{ (G \abs{\mb{x}}) }_2$.
It is straightforward to modify the algorithms of~\sctn{algl1inf} for this problem, by replacing the proximal operator of the max function by the proximal operator of the $\ell_2$ norm~\cite[Sec. 6.5.1]{parikh-2014-proximal}.

\section{Results}
\label{sec:results}

\begin{figure*}[htbp]
  \centering \small
  \begin{tabular}{ccccc}
    \subfigure[\label{fig:tstimg0}\protect\rule{0pt}{1.5em}
    Image 1]
               {\includegraphics[width=3.2cm]{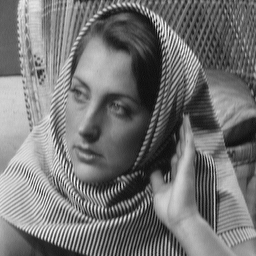}} &
    \hspace{-2mm}
    \subfigure[\label{fig:tstimg1}\protect\rule{0pt}{1.5em}
    Image 2]
               {\includegraphics[width=3.2cm]{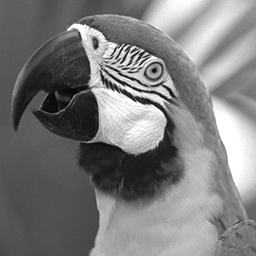}}
&
    \hspace{-2mm}
    \subfigure[\label{fig:tstimg2}\protect\rule{0pt}{1.5em}
    Image 3]
               {\includegraphics[width=3.2cm]{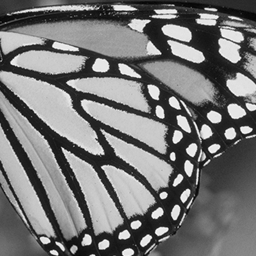}}
&
    \hspace{-2mm}
    \subfigure[\label{fig:tstimg3}\protect\rule{0pt}{1.5em}
    Image 4]
               {\includegraphics[width=3.2cm]{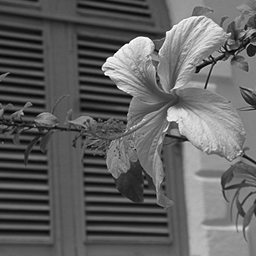}}
&
    \hspace{-2mm}
    \subfigure[\label{fig:tstimg4}\protect\rule{0pt}{1.5em}
    Image 5]
               {\includegraphics[width=3.2cm]{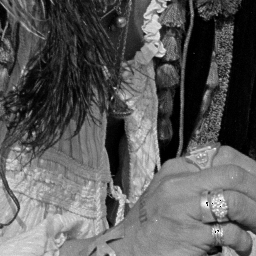}}
  \end{tabular}
  \vspace{-3mm}
  \caption{Set of $256 \times 256$ pixel noise-free test images.}
  \label{fig:tstimg}
\end{figure*}

We assess problems~\eq{cbpdnl1infty} and~\eq{cbpdnl12} by comparing their performance with that of problem~\eq{convbpdn} in a Gaussian white noise denoising problem. The same convolutional dictionary, consisting of 128 filters of size $8 \times 8$ samples, learned from a set of ten training images (selected from images on Flickr with a Creative Commons license) of $1024 \times 1024$ pixels each, was used in all cases.

A set of five greyscale reference images, depicted in~\fig{tstimg}, was constructed by cropping regions of $256 \times 256$ pixels from well-known standard test images. The regions were chosen to contain diversity of content while avoiding large smooth areas, and the size was chosen to be relatively small so that it would be computationally feasible to optimise method parameters via a grid search. The reference images were scaled so that pixel values were in the interval $[0,1]$, and corresponding test images were constructed by adding Gaussian white noise with a standard deviation of 0.05 (a relatively mild noise level).

In all cases the CSC-based denoising was achieved as follows: lowpass filter the noisy image\footnote{The lowpass filtered signal was computed by Tikhonov regularization with a gradient term~\cite[pg. 3]{wohlberg-2017-sporco}, with regularization parameter $\lambda = 2.0$.}, sparse code the highpass residual~\cite[Sec. I]{wohlberg-2016-efficient}, and reconstruct the image from the sparse representation and add back the lowpass component to obtain the final denoised image. The denoising performance of each methods was individually optimised for each image via a search over a logarithmically spaced grid on the $\lambda$ parameter. For CSC $\ell_1$, the penalty parameter $\rho$ was set automatically using the residual balancing strategy~\cite[Sec. 3.4.1]{boyd-2010-distributed}\cite{wohlberg-2015-adaptive}.
Since this strategy is ineffective for the mixed group norm forms of CSC, the penalty parameters for these methods were set, based on numerical experiments, to $0.05 \lambda$ and $3.0 \lambda$ for CSC $\ell_{1,\infty}$ and CSC $\ell_{1,2}$ respectively. Similarly, the value of $\alpha_0$ was set to $0.06$ and $0.03$ for CSC $\ell_{1,\infty}$ and CSC $\ell_{1,2}$ respectively. In both cases the setting $\alpha_1 = \alpha_0^{-1}$ was used. In order to ensure convergence, 250 iterations were allowed for the CSC $\ell_1$ algorithm, and 350 iterations were allowed for the mixed norm algorithms.

For comparison purposes, results were also computed using a standard patch-based denoising scheme that is essentially the same as that used in the well-known K-SVD denoising technique~\cite{elad-2006-image}, with a $64 \times 128$ dictionary learned from a separate image training set (the same set used to learn the convolutional dictionary described above). These results are labelled ``OMP'' in the tables of results.

{\small
\begin{table}[htbp]
  \centering
  \begin{tabular}{|l|r|r|r|r|r|} \hline
   & \multicolumn{5}{|c|}{Test Image} \\ \hline
Method       &  1     & 2     & 3     & 4     & 5   \\ \hline\hline
OMP          & \textbf{30.38}  & \textbf{33.31} & \textbf{30.47} &  \textbf{32.40} & \textbf{30.54} \\ \hline\hline
CSC $\ell_1$ & 29.12 & 32.76 & 29.64 & 31.21 & 29.92  \\ \hline
CSC $\ell_{1,\infty}$ & 26.19 & 27.52 & 27.43 & 29.50 & 28.81  \\ \hline
CSC $\ell_{1,2}$ & 28.19 & 30.45 & 29.03 & 30.67 & 29.69  \\ \hline
 \end{tabular} \vspace{1mm}
 \caption{Comparison of denoising performance (PSNR in dB) of the different denoising methods for each of the five test images corrupted by Gaussian white noise with $\sigma = 0.05$.
Bold values indicate the best performing method.}
  \label{tab:cmp1}
\end{table}
}

\begin{figure}[htbp]
  \centering \small
  \begin{tabular}{cc}
    \hspace{-2mm}
    \subfigure[\label{fig:cbpdnartifact0}\protect\rule{0pt}{1.5em}
    Noisy]
               {\includegraphics[width=4.1cm]{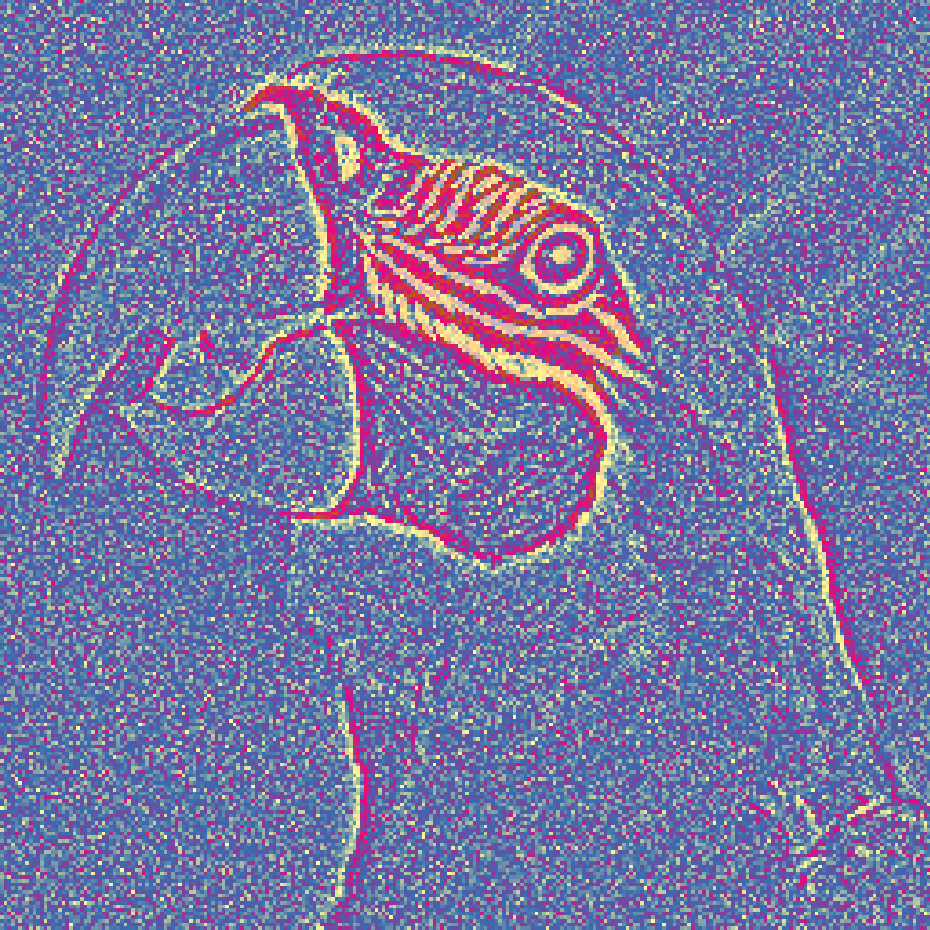}} &
    \hspace{-2mm}
    \subfigure[\label{fig:cbpdnartifact1}\protect\rule{0pt}{1.5em}
    CSC $\ell_1$ denoised]
               {\includegraphics[width=4.1cm]{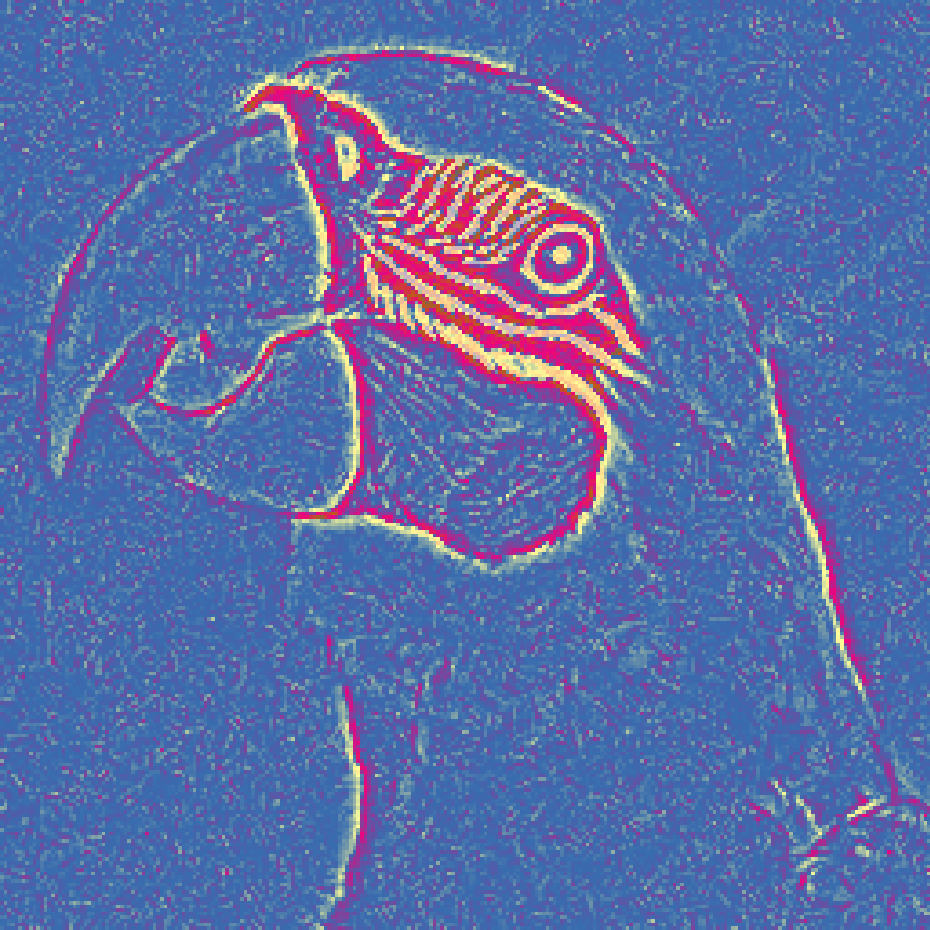}}
  \end{tabular}
  \vspace{-3mm}
  \caption{Highpass components of noisy test image and corresponding CSC $\ell_1$ denoised image illustrating the artifacts resulting from this denoising: faint ``ghosts'' of dictionary filters are visible where they have non-negligible correlation with the local noise pattern. Colour map selected to enhance visibility of the artifacts. (For best visibility, this figure should be viewed zoomed-in in the electronic version of the document.)}
  \label{fig:cbpdnartifact}
\end{figure}

The denoising performance of these methods is compared in~\tab{cmp1}.  It is immediately apparent that the $\ell_{1,\infty}$ penalty gives very much worse denoising performance than the standard $\ell_1$ penalty, and while the performance of the $\ell_{1,2}$ penalty is somewhat better, it is also substantially inferior to that of the $\ell_1$ penalty. To put these results in context, note that the performance of the usual CSC $\ell_1$ is itself quite poor when compared with the patch-based ``OMP'' results. Aside from the poor PSNR, the CSC $\ell_1$ method also exhibits faint filter ``ghost'' artifacts, as illustrated in~\fig{cbpdnartifact}. (These artifacts are also encountered in patch-based denoising methods, but the patch aggregation by averaging is very effective in suppressing them in the final image.)

\begin{figure}[htbp]
  \centering \small
  \includegraphics[width=7.7cm]{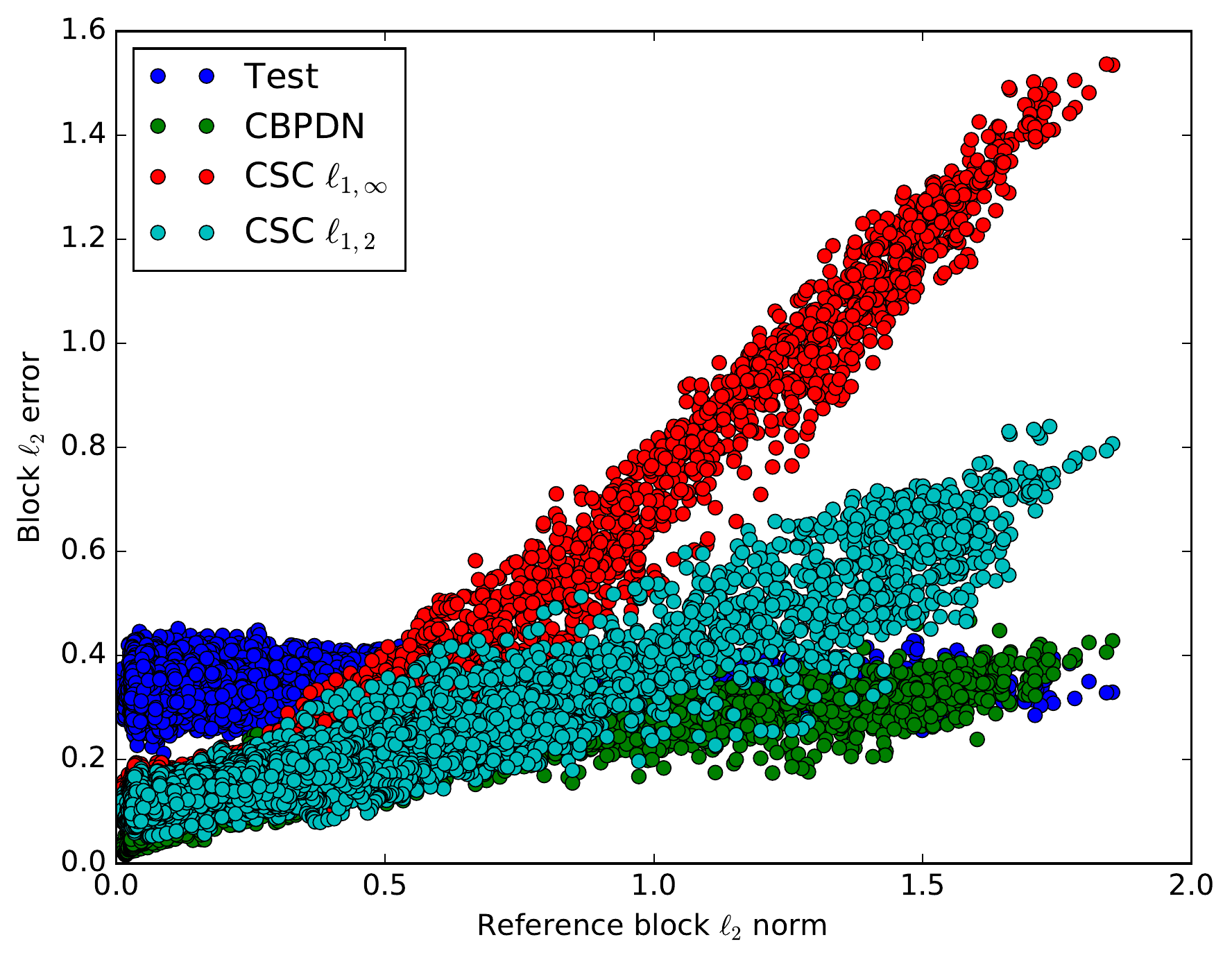}
  \vspace{-2mm}
  \caption{Scatter plot of image block error against the norm of the corresponding block in the reference image, for blocks extracted from the noisy test image and the different CSC denoised images.
}
  \label{fig:blkerr1}
\end{figure}

To understand the poor performance of CSC $\ell_{1,\infty}$, we look at the $\ell_2$ errors of individual image blocks (of the same size as the dictionary filters) in the highpass filtered images, plotted against the $\ell_2$ norm of the corresponding blocks in the highpass filtered reference Image 2 (see~\fig{tstimg}). The reason is immediately clear from~\fig{blkerr1}: low-contrast/smooth image regions are not sufficiently  regularized, and high-contrast/edge regions are greatly over-regularized\footnote{The results in~\fig{blkerr1} are for the minimum MSE choice of $\lambda$. If $\lambda$ is larger the over-regularization of large norm blocks is even worse, and when it is smaller there is negligible regularization of low norm blocks.}. This is due to a basic property of regularization with the $\ell_{\infty}$ norm, which effectively selects some threshold above which values are shrunk to that threshold, and below which they are unaffected~\cite[pg. 9]{bach-2012-optimization}. In the context of a mixed $\ell_{1,\infty}$ group norm, this implies that all groups will have the same $\ell_1$ norm: either low contrast regions are not sparse enough, or high contrast regions are too sparse.
The  $\ell_{1,2}$ norm exhibits similar but much less severe behaviour.

\section{Group Weighting}
\label{sec:weighting}

There are two obvious approaches to dealing with the weakness of the $\ell_{1,q}$ group norms applied to images with varying local contrast: apply some form of local contrast normalization to the image to be represented, or apply suitable weighting factors to the mixed norms to compensate for the varying sparsity requirements. We consider the latter approach here due to the substantially greater complexity for the former\footnote{Contrast normalization is often applied when CSC is used for classification tasks~\cite{zeiler-2011-adaptive}, but those methods are usually not appropriate for reconstruction problems. Furthermore, the overlapping nature of the groups makes it difficult to choose a suitable normalization factor based on the contrast of the spatial region corresponding to each group.}.

We consider two different types of weighting. The first of these replaces the outer $\ell_q$ norm in the mixed $\ell_{1,q}$ norm with a weighted $\ell_q$ norm. The two specific cases we consider here become $\norm{\mb{x}}_{1,\infty}  = \max_i \parensz[\big]{ w_i \norm{ g_i(\mb{x}) }_1 }$ and $\norm{\mb{x}}_{1,2} = \sqrt{ \sum_i w_i \norm{ g_i(\mb{x}) }_1^2 }$, where the $w_i$ are distinct weights for each group. Modifying the algorithms in~\sctn{algl1inf} and~\sctn{l12} to use these weighted mixed norms requires closed forms for the proximal operators of the weighted max function and weighted $\ell_2$ norm, which are straightforward to derive following the same approaches used for the corresponding unweighted norms (Sec. 6.4.1, 6.5.2, and 6.5.1 in ~\cite{parikh-2014-proximal}). The second type involves replacing the inner $\ell_1$ norm with a weighted $\ell_1$ norm, which is easily achieved by replacing the unit kernels defining operator $G$ (see~\sctn{algl1inf}) with a kernel consisting of the desired weights.

The natural choice of group weighting $w_i$, given the results in~\fig{blkerr1}, is to make it inversely proportional to a measure of image activity in the spatial region corresponding to each group $i$, so that more active regions are penalised less, and vice versa. A variety of weight construction schemes were empirically compared, the most effective of which was to use the local squared $\ell_2$ norm of the image region corresponding to each coefficient group (which is easily computed by convolving the squared coefficient maps with a appropriately sized kernel of unit entries) as the activity measure. The final group weights $w_i$ were obtained by taking the inverse of the sum of the activity measure across all coefficient maps at each spatial location.

The weightings for the inner $\ell_1$ norm (\ie the kernel defining operator $G$) were defined by observing that, while non-convolutional dictionaries are usually normalised so that each column has unit norm, the ``stripe dictionary'' depicted in~\fig{dctptch} cannot be so normalised due to its construction from translations of the generating block dictionary. This lack of normalisation can be compensated for, however, if the $\ell_1$ norm of a group is appropriately weighted, taking into account the norm of each translated part of the stripe dictionary.

{\small
\begin{table}[htbp]
  \centering
  \begin{tabular}{|l|r|r|r|r|r|} \hline
   & \multicolumn{5}{|c|}{Test Image} \\ \hline
Method       &  1     & 2     & 3     & 4     & 5   \\ \hline\hline
OMP          & \textbf{30.38}  & 33.31 & \textbf{30.47} &  \textbf{32.40} & 30.54 \\ \hline\hline
CSC $\ell_1$ & 30.22 & \textbf{33.39} & 30.28 & 31.93 & \textbf{30.56}  \\ \hline
CSC $\ell_{1,\infty}$ & 29.22 & 32.07 & 29.54 & 30.91 & 29.85 \\ \hline
CSC $\ell_{1,2}$ & 29.26 & 32.76 & 29.77 & 31.15 & 30.04 \\ \hline
 \end{tabular} \vspace{1mm}
 \caption{Comparison of denoising performance (PSNR in dB) of the different weighted denoising methods for each of the five test images corrupted by Gaussian white noise with $\sigma = 0.05$.
Bold values indicate the best performing method.}
  \label{tab:cmp2}
\end{table}
}

\begin{figure}[htbp]
  \centering \small
  \includegraphics[width=7.7cm]{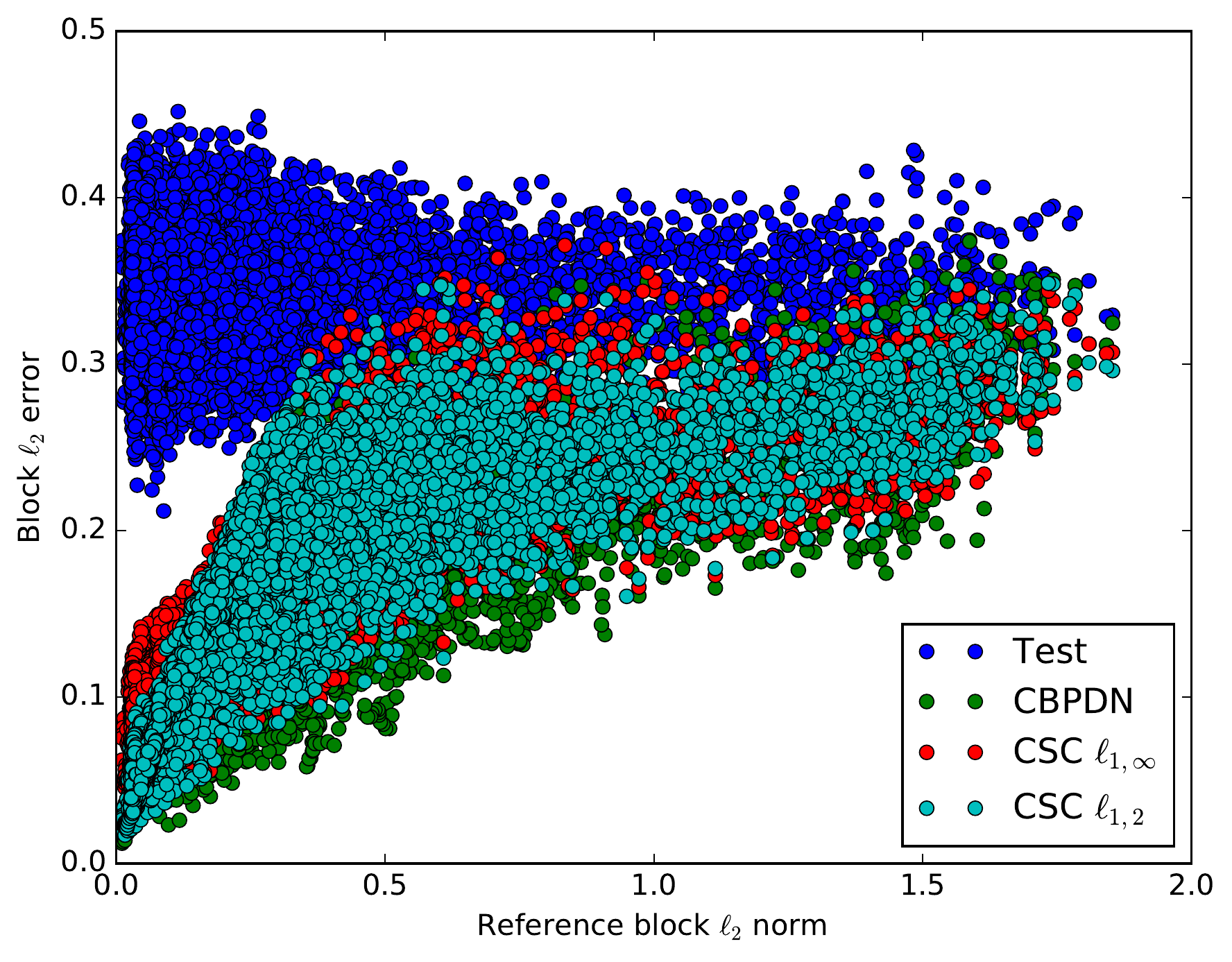}
  \vspace{-2mm}
  \caption{Scatter plot of image block error against the norm of the corresponding block in the reference image, for blocks extracted from the noisy test image and the different CSC denoised images.
}
  \label{fig:blkerr2}
\end{figure}

The results obtained using this weighting scheme for CSC $\ell_{1,\infty}$ and CSC $\ell_{1,2}$ are displayed in~\tab{cmp2}. By comparing with~\tab{cmp1}, it is apparent that the weighting greatly improves the performance of these methods, making them competitive with the CSC $\ell_1$ results reported in ~\tab{cmp1}. The substantial performance improvement is also apparent from the block error plots displayed in~\fig{blkerr2}.

\section{Weighting the $\ell_1$ Norm}
\label{sec:l1weighting}

To provide a fair comparison, a variety of weighting schemes for the $\ell_1$ norm were also investigated. Of these, the most effective was computed as the inverse of $(D^T \mb{s})^2$, where $D$ is as defined in~\eq{dxcbpdn} and $\mb{s}$ is the highpass component of the noisy test image. ($D^T \mb{s}$ can computed in the DFT domain by multiplying $\hat{\mb{s}}$ by the complex conjugate of the dictionary filters $\hat{\mb{d}}_m$.) It can be seen from~\tab{cmp2} that this weighting scheme significantly improves the PSNR, making it competitive with that of the patch-based method ``OMP''. It is also effective in suppressing the filter ``ghost'' artifacts discussed in~\sctn{results}, which is not surprising considering that the direct effect of the weighting scheme is to penalise filters that are not locally correlated with the signal.

\section{Conclusions}
\label{sec:cnclsn}

We have proposed the first algorithms for solving the difficult optimization problems corresponding to CSC with $\ell_{1,q}$ group norms with overlapping groups, enabling an empirical examination of the properties of this form of CSC.
It is interesting to observe that, although a recent theoretical analysis of convolutional sparse representations suggests that the $\ell_{1,\infty}$ might have favourable performance in practical applications~\cite[Sec. VI]{papayan-2017-working2}, the experiments presented here  indicate that it is greatly inferior to the $\ell_1$ norm for denoising of images subject to Gaussian white noise. The $\ell_{1,2}$ norm is superior to the $\ell_{1,\infty}$ norm, but also not competitive with the $\ell_1$ norm.

The introduction of suitable weighting schemes greatly improves the performance of the $\ell_{1,\infty}$ and $\ell_{1,2}$ norms, shrinking the performance gap between them, and making them competitive with the unweighted $\ell_1$ norm. However, the performance of the $\ell_1$ norm can also be improved by appropriate weighting, and the resulting method is again superior to the weighted $\ell_{1,q}$ norms. It is particularly noteworthy that this improvement makes CSC with the $\ell_1$ norm competitive with the more common patch-based sparse denoising methods, a result which has not previously been reported in the literature.

Although the performance of CSC with the $\ell_{1,q}$ group norms in the denoising problem is disappointing, it is hoped that the demonstration that these problems are computationally expensive but tractable will spur further research, including the examination of alternative applications for which they may be more appropriate, and the development of more effective weighting schemes to improve their performance. Implementations of the algorithms proposed here will be included in a future release of the SPORCO library~\cite{wohlberg-2016-sporco, wohlberg-2017-sporco} as an aid to the reproducibility of this research.

\section{Acknowledgment}

The author thanks Wotao Yin and Alessandro Foi for valuable discussions regarding various aspects of this work.

\bibliographystyle{IEEEtranD}
\bibliography{strcsprs}

\end{document}